# Exploring Data Augmentation Methods on Social Media Corpora

**Isabel Garcia Pietri[1], Kineret Stanley[2]**

*University of California, Berkeley*
*School of Information*

[1]isabelgarpietri@berkeley.edu, [2]kbaumberg@berkeley.edu

## Abstract

Data augmentation has proven widely effective in computer vision. In Natural Language Processing (NLP) data augmentation remains an area of active research. There is no widely accepted augmentation technique that works well across tasks and model architectures. In this paper we explore data augmentation techniques in the context of text classification using two social media datasets. We explore popular varieties of data augmentation, starting with oversampling, Easy Data Augmentation (Wei and Zou, 2019) and Back-Translation (Sennrich et al., 2015). We also consider Greyscaling, a relatively unexplored data augmentation technique that seeks to mitigate the intensity of adjectives in examples. Finally, we consider a few-shot learning approach: Pattern-Exploiting Training (PET) (Schick et al., 2020). For the experiments we use a BERT transformer architecture.

Results show that augmentation techniques provide only minimal and inconsistent improvements. Synonym replacement provided evidence of some performance improvement and adjective scales with Grayscaling is an area where further exploration would be valuable. Few-shot learning experiments show consistent improvement over supervised training, and seem very promising when classes are easily separable but further exploration would be valuable.

## 1. Introduction

Small size datasets impose many challenges for deep learning approaches. Neural networks architectures, such as convolutional neural networks, require many examples to achieve good performance. On the other hand, fine tuning pre-trained models such as BERT using small datasets results in a high variance in performance. Even with the same hyperparameter values, distinct random seeds can lead to substantially different results (Dodge et al., 2020)

In this study we explore data augmentation techniques for NLP in the context of text classification. Data augmentation ("DA") is the process of creating synthetic data based on existing training data, and seeks to reduce overfitting. In NLP, where the input space is discrete, how to generate effective augmented examples remains an area of active research.

We explore various data augmentation techniques. We start with a simple oversampling, and move towards techniques more appropriate for NLP. Among data augmentation techniques, we select Easy Data Augmentation (Wei and Zou, 2019) and Back-Translation (Sennrich et al., 2015). We also explore Greyscaling adjectives as suggested in a review of data augmentation techniques (Feng et al., 2021) and (Soler et al., 2020). Additionally we consider Pattern-Exploiting Training (PET), as a few-shot learning approach (Schick et al., 2020).

## 2. Related Work

Several studies have explored data augmentation techniques in NLP. Some techniques are based on token-level perturbations including random replacement, insertion, deletion and swap (Wei and Zou, 2019). Language models have also been used for DA. The popular back-translation method (Sennrich et al., 2015) translates a sequence into another language and then back into the original language. Pretrained language models such as RNNs (Kobayashi, 2018) and transformers (Yang et al., 2020) have also been used for augmentation.

In computer vision, Greyscaling (toning down the intensity) reflects differences in perspective. In NLP, replacing adjectives with ranked scalar adjectives (Soler et al., 2020) uses synonyms ranked on a scale as an augmentation technique.

Few-shot learning approaches are popular in scarce labeled data scenarios as well. Approaches for few-shot learning in NLP include exploiting examples from related tasks (Yu et al., 2018; Yin et al., 2019) using data augmentation (Xie et al., 2020; Chen et al., 2020), and exploiting patterns to reformulate input examples (Schick et al., 2020).

## 3. Data

In this study we use two social media datasets: TRAC-2 and RtGender datasets.

### 3.1 Aggression Identification in Social Media dataset (TRAC-2)

This dataset comes from the Second Workshop on Trolling, Aggression and Cyberbullying (TRAC-2) (Kumar et al., 2020). The dataset originally contains data in three languages (Bengali, Hindi and English) but we only consider the data in English; which, consists of ~6k comments from YouTube labeled for two tasks:

**Task A-** identify aggression: Non-aggressive (NAG), Covertly Aggressive (CAG) and Overtly Aggressive (OAG).

**Task B-** gendered aggression identification: Non-gendered (NGEN) and Gendered (GEN).

The dataset is highly imbalanced. The table 1 shows the data split and class distribution.

|  | Total | Task A (%) | | | Task B (%) | |
| --- | --- | --- | --- | --- | --- | --- |
|  |  | NAG | CAG | OAG | NGEN | GEN |
| Train | 4,263 | 79.2 | 10.6 | 10.2 | 92.8 | 7.2 |
| Development | 1,066 | 78.4 | 11.0 | 10.6 | 93.2 | 6.8 |
| Test | 1,200 | 57.5 | 18.7 | 23.8 | 85.4 | 14.6 |

TABLE 1. TRAINING, DEVELOPMENT AND TEST DATASETS SIZES AND LABEL DISTRIBUTION (TRAC-2)

### 3.2 RtGender dataset

Corpus (Voigt et al., 2018) provided by Stanford, containing social media posts from several platforms (Reddit, Facebook, Ted talk comments), where the gender of the poster and respondents are labeled. We focused on an annotated subset of response text (~15k total). These data are labeled for sentiment (positive, neutral, mixed, negative) and relevance of the responses "to the source." We split our data for train (70%), development (15%), and test (15%). The dataset is highly imbalanced with almost half of the comments labeled positive. We run experiments on the 1) entire dataset, and 2) a subset with three classes (excluding sentiment is 'mixed') and excluding examples labeled as 'irrelevant' (~12% of the total dataset). See Appendix C. for more details about examples labeled as 'mixed.' The table 2 shows the data split and class distribution.

|  | Total | Sentiment (%) | | | |
| --- | --- | --- | --- | --- | --- |
|  |  | Pos. | Neut. | Mi. | Neg. |
| Train | 10,746 | 49.3 | 24.3 | 10.1 | 16.3 |
| Development | 2,303 | 48.5 | 24.6 | 10.3 | 16.5 |
| Test | 2,303 | 47.5 | 24.8 | 17.7 | 10.0 |

TABLE 2. TRAINING, DEVELOPMENT AND TEST DATASETS SIZES AND LABEL DISTRIBUTION (RTGENDER)

## 4. Methodology

During the first part of this study, we focus on building classification models using only the data available in the datasets. Below we describe the architecture of the models we consider.

### 4.1 Neural bag of words models (BOW)

This is our baseline model. The model architecture is very simple: average of the input word embeddings, a single dense layer, a dropout layer and an output layer. We use 300-dimensional static word embeddings from Global Vectors (GloVe) (Pennington et al., 2014). No fine-tuning of the word embeddings is allowed during training. To convert the sentences into tokens we use the tokenizer available in Keras.

### 4.2 BERT

Bidirectional Encoder Representation from Transformers (BERT) models (Devlin et al., 2019). We consider the BERT Base model uncased. To build the models, we use the open source transformers library by HuggingFace Inc. (Wolf et al., 2019)[1]. Each model is trained five times using different seeds and average results are reported.

### 4.2.2 Data Augmentation

In this study we explore several data augmentation methods; which, aim to prevent overfitting using various augmentation techniques to create synthetic data.

---

[1] Training procedure and hyperparameters information is available on the appendix section.

**Oversampling (O):** One of the simplest ways to create synthetic data is to generate exact copies of the available data. We do this by sampling minority class data.

**Easy Data Augmentation (EDA):** We use data augmentation techniques proposed in the paper Easy Data Augmentation (Wei and Zou, 2019). We experiment with two of the methods presented in the paper: synonym replacement and random deletion. Synonym replacement randomly chooses *n* words from the sentence that are not stop words, and replaces each of these words with one of its synonyms chosen at random. Random deletion randomly removes words in the sentence.

**Back-Translation (BT):** We use the back-translation method to create variations of the original training data while preserving the context. We translate the training data to other languages and then translate it back to english. We use the MarianMT model available in Hugging Face, which is based on translation models trained on the OPEN Parallel Corpus (OPUS) using MarianNMT translation framework (Junczys-Dowmunt et al., 2018).

**Greyscaling- Using Adjective Scales to Tone Down Sentiment:** To generate augmented examples, we started with three scalar adjective data sets. The word scales are not mutually exclusive.

1. de Melo (de Melo et al., 2013) which used Mixed Linear Programming to jointly determine intensity order information. The scales (e.g., horrible to awesome) are divided into half-scales (e.g., horrible to bad). We used 87 half-scales.

2. Crowd (Cocos et al., 2018) adjective scales sourced from the Paraphrase Database; also, represented with half-scales. We used 79 half-scales.

3. Wilkinson (Wilkinson et al., 2016) adjectives sourced through crowdsourcing. We used 21 half-scales.

First, we augment examples using the three adjective scales, with all milder options returned. ['harmful', 'toxic', 'deadly'] →{deadly: ['toxic', 'harmful'], toxic: ['harmful']}

Then we run our augmented dataset through the same BERT configuration as our base BERT case.

**Greyscaling with contextual representations:** To ensure that the synthetic examples are appropriate in context (table 7 for examples), we perform a second experiment with the Greyscale augmented datasets for RtGender and TRAC-2 using the Universal Sentence Decoder, with the transformer approach (Cer et al., 2018). We employ the same similarity calculation computing the cosine similarity and converting it into an angular distance. This score indicates how similar the context of our augmented and original examples are.

$$sim(u, v) = 1 - arccos\ u \cdot v\ ||u||\ ||v||\ /\pi$$

Finally, we remove examples from the bottom quartile of scores and run the remaining augmented Greyscaled examples through BERT.

### 4.2.2.1 Augmentation procedure

We first oversample all datasets to achieve uniform distribution of classes. We then use the oversampled datasets for the remaining augmentation experiments.

To augment using the EDA methodology we create 3 extra examples per original example in the training data. Similarly, for the back-translation augmentation we translate to Spanish, French and Italian to obtain 3 extra examples. For Greyscaling, augmentation was specific to each example, resulting in augmented datasets of ~2x the size.

Table 3 shows some examples of augmented text in the TRAC-2 dataset.

| **Who is she.....may be she has escaped from some mental hospital...** | |
|---|---|
| (1) BT | Who are you? Maybe she ran away from a psychiatric hospital. |
| (2) BT | She may have escaped from a psychiatric hospital. |
| (3) BT | Who is she... maybe she escaped from a psychiatric hospital... |
| (1) EDA | who is she may be she has escaped from some genial hospital |
| (2) EDA | who is she english hawthorn be she has escaped from some mental hospital |
| (3) EDA | who is she may be she has break away from some mental hospital |
| (1) GS | Who is she.....may be she has escaped from few mental hospital... |

TABLE 3. AUGMENTATION EXAMPLES USING BACK-TRANSLATION(BT) EASY DATA AUGMENTATION (EDA) AND GREYSCALING (GS) TECHNIQUES

### 4.2.3 Few-shot learning: Pattern-Exploiting Training (PET)

In the traditional supervised training in NLP, we give examples and corresponding labels to a model and we expect the model to learn the underlying task. This generally works well when we have a big amount of data but it is increasingly difficult in a few-shot learning escenario.

We explore PET (Pattern-Exploiting Training), an approach for few-shot learning in NLP (Schick et al., 2020). This is a semi-supervised training procedure that reformulates input examples as close questions to help language models to understand a given task.

The PET methodology works in three steps (Figure 1):
1. A number of patterns are created to convert training examples to close questions. For each pattern a separate pre-trained language model is finetuned on a small training set $T$.
2. The ensemble of all models is then used to annotate a large unlabeled dataset D with soft labels.
3. Finally, a standard classifier is trained on the soft-labeled dataset.

There is another variant of the process called iPET, which is an iterative variant of PET, where steps 1 and 2 are repeated with increasing training set sizes.

We evaluate the PET and iPET methodologies on the Task-A of the TRAC-2 dataset and on the full RtGender dataset. We work with two versions of the Task-A aggression detection classification problem: a simplification considering a binary classification problem (we combine the Covertly Aggressive and Overtly Aggressive classes) and the original 3-classes classification problem.

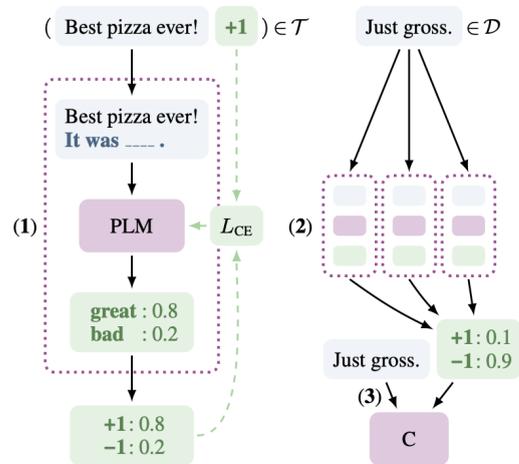

Fig. 1 PET for sentiment classification. (Source: https://arxiv.org/abs/2001.07676)

For all experiments we use BERT base uncased as language model. We investigate the performance of PET and iPET for different training set sizes. Each final classifier is trained five times using different seeds and average results are reported. We use the available defaults for all other hyperparameters.

To create the training sets $T$, we sample without replacement the original training set, sampling the same amount of examples for each class. The unused training examples are considered unlabeled (set $D$). For instance, in the TRAC-2 dataset to create a training set $T$=100, we sample without replacement 100 across the classes, and the remaining 4,163 examples are considered the unlabeled set $D$.

**Patterns and verbalizers TRAC-2 dataset (Task-A with 2 classes):** Patterns are the encoding functions to convert the input text to close questions, and verbalizers are the words (tokens) that replace the masks in the patterns.

We define the following patterns for an input text $x$:
$P_1(x) = x$. In summary, this is _ _ _
$P_2(x) = x$. You are just _ _ _ !
$P_3(x) = $ It was _ _ _. $x$.
$P_4(x) = $ Just _ _ _ ! $x$.

We hypothesize that aggressive comments would have a higher probability to be related with the word *bad*, and non-aggressive comments would have a higher probability to be related with the word *good* (verbalizers: *good* and *bad*).

**Patterns and verbalizers TRAC-2 dataset (Task-A with 3 classes):** The patterns are the same and we use the following verbalizers: *good* (for non-aggressive), *bad* (for covertly aggressive), *terrible* (for overtly aggressive).

**Patterns and verbalizers RtGender dataset:** The patterns remain the same. We use the following verbalizers: *positive*, *neutral*, *mixed* and *negative*.

## 4. Results

The tables 4, 5 and 6 summarize the results obtained from all experiments.

As a reference, the best systems submitted in the TRAC-2 workshop reported a weighted f1-score of approximately 0.80 for Task-A and 0.87 for Task-B (systems were ensembles of models).

| Model + Augmentation | Task-A | | | Task-B | | |
|---|---|---|---|---|---|---|
| | F1 - macro | F1-weig. | Acc | F1-macro | F1-weig. | Acc |
| BoW | 0.505 ±0.056 | 0.626 ±0.04 | 0.693 ±0.024 | 0.602 ±0.129 | 0.83 ±0.04 | 0.864 ±0.01 |
| BERT | 0.562 ±0.068 | 0.671 ±0.051 | 0.716 ±0.033 | **0.721** ±0.009 | **0.856** ±0.008 | **0.852** ±0.014 |
| BERT + Over-sampling (O) | 0.604 ±0.027 | 0.696 ± 0.02 | 0.722 ±0.021 | 0.698 ±0.017 | 0.836 ±0.029 | 0.825 ±0.04 |
| BERT + EDA (+ O) | **0.618** ±0.056 | **0.706** ±0.041 | **0.727** ±0.025 | 0.698 ±0.016 | 0.843 ±0.022 | 0.837 ±0.034 |
| BERT + Back transl. (+O) | 0.587 ±0.033 | 0.681 ±0.03 | 0.706 ±0.024 | 0.699 ±0.018 | 0.845 ±0.01 | 0.84 ±0.014 |
| BERT + Greysc. (+ O) | 0.607 ±0.031 | 0.694 ±0.024 | 0.719 ±0.017 | 0.715 ±0.015 | 0.849 ±0.016 | 0.841 ±0.024 |
| BERT + Greysc. + C.R. (+ O) | 0.616 ±0.013 | 0.704 ±0.013 | 0.723 ±0.013 | 0.706 ±0.013 | 0.846 ±0.007 | 0.838 ±0.01 |
| BERT w/ PET[2] | -- | -- | 0.706 ±0.003 | -- | -- | -- |

TABLE 4. CLASSIFICATION MODELS FOR AGGRESSION IDENTIFICATION (TRAC-2). METRICS AVERAGE AND STANDARD DEVIATION

| Model + Augmentation | Overall | | | excluding Irrelevant & Mixed Examples[3] | | |
|---|---|---|---|---|---|---|
| | F1 - macro | F1-weig. | Acc | F1 - macro | F1-weig. | Acc |
| BoW | 0.431 ±0.005 | 0.553 ±0.004 | 0.693 ±0.024 | 0.613 ±0.014 | 0.688 ±0.01 | 0.69 ±0.08 |
| BERT | 0.555 ± 0.006 | 0.66 ± 0.008 | 0.665 ± 0.01 | 0.700 ± 0.004 | 0.764 ± 0.002 | **0.771** ± 0.002 |
| BERT + Over-sampling (O) | **0.56** ±0.013 | **0.665** ± 0.01 | **0.671** ±0.006 | 0.693 ± 0.003 | 0.76 ± 0.003 | 0.762 ± 0.005 |
| BERT + EDA (+ O) | 0.536 ±0.01 | 0.642 ±0.006 | 0.65 ±0.005 | 0.689 ±0.015 | 0.754 ± 0.011 | 0.756 ± 0.006 |
| BERT + Back transl. (+ O) | 0.527 ±0.007 | 0.633 ±0.004 | 0.648 ±0.005 | 0.688 ±0.007 | 0.754 ± 0.006 | 0.759 ± 0.007 |
| BERT + Greysc. (+ O) | 0.541 ±0.011 | 0.642 ±0.001 | 0.634 ±0.015 | 0.682 ±0.012 | 0.748 ±0.009 | 0.751 ±0.012 |
| BERT + Greysc. + C.R. (+ O) | 0.549 ±0.008 | 0.655 ±0.007 | 0.652 ±0.011 | **0.704** ±0.008 | **0.765** ±0.007 | 0.766 ±0.011 |
| BERT w/ PET[4] | -- | -- | 0.588 ±0.003 | -- | -- | -- |

TABLE 5. CLASSIFICATION MODELS FOR RTGENDER SENTIMENT IDENT. METRICS AVERAGE AND STANDARD DEVIATION

| T | Method | TRAC-2 (Task-A) 2 classes[5] | TRAC-2 (Task-A) 3 classes | RtGender |
|---|---|---|---|---|
| 10 | supervised | 0.437 ±0.129 | 0.298 ±0.135 | 0.280 ±0.060 |
| | PET | 0.711 ±0.007 | 0.612 ±0.005 | 0.179 ±0.002 |
| | iPET | 0.825 ±0.002 | 0.642 ±0.003 | 0.147 ±0.003 |
| 50 | supervised | 0.655 ±0.123 | 0.436 ±0.179 | 0.308 ±0.111 |
| | PET | 0.818 ±0.002 | 0.623 ±0.004 | 0.336 ±0.002 |
| | iPET | 0.833 ±0.002 | 0.619 ±0.004 | 0.347 ±0.003 |
| 100 | supervised | 0.770 ±0.027 | 0.385 ±0.165 | 0.333 ±0.109 |
| | PET | 0.839 ±0.003 | 0.643 ±0.004 | 0.493 ±0.002 |
| | iPET | 0.833 ±0.002 | 0.642 ±0.004 | 0.497 ±0.002 |
| 500 | supervised | 0.834 ±0.009 | 0.677 ±0.024 | 0.525 ±0.030 |
| | PET | 0.850 ±0.002 | 0.706 ±0.003 | 0.573 ±0.003 |
| | iPET | 0.844 ±0.004 | 0.705 ±0.007 | 0.588 ±0.003 |

TABLE 6. PET and iPET EXPERIMENTS WITH VARIOUS NUMBER OF LABELED EXAMPLES. MEAN ACCURACY AND STANDARD DEVIATION

---

[2] Best accuracy was for PET for 500 labeled samples for TRAC-2 Task A (3-classes).

[3] Ran iterations of 3 instead of 5 for this subset.
[4] Best accuracy was for iPET for 500 labeled samples for RtGender.
[5] Accuracy of the model trained with the full dataset in the 2-classes classification problem is 0.838 ±0.057.

## 5. Discussion

We observe that oversampling offers a minimal or no performance improvement across tasks. Given that the classes are initially very imbalanced, we expected a higher performance improvement. We attribute this to overfitting. The distribution of augmented data is identical to the original data, leading to greater overfitting.

EDA techniques offered minimal performance improvement for TRAC-2 Task-A and negligible (or no) improvements in other tasks. This is consistent with the previous research ([Longpre et al., 2020](#)) that shows that these general augmentation methods do not benefit modern Transformer architectures.

Back-translation augmentation provided negligible or no performance improvements across tasks. This method generated augmented text with some variations without losing meaning. The selected romance languages (Spanish, Italian, French) have the advantage of developed language models and easy translation into English. As we can observe in [table 3](#), some augmentations using EDA lose meaning but resulted in better performance for TRAC-2 Task-A. This could indicate that diversity is more important than quality, as found in some previous research ([Xie et al., 2020](#)).

Similarly, Greyscaling augmentation provided negligible or no performance improvements across tasks. But, once we added in contextual representations the overall scores aligned well with those of our EDA experiment for TRAC-2 Task-A.

Few-shot learning methodology PET worked well in problems where classes are easily separable. As we observe in [table 6](#), for the 3-classes classification problem TRAC-2 Task-A, none of the PET/iPET models achieve the accuracy levels we obtain when we train with all the training data (also the case of RtGender). However, in the 2-classes scenario of the same problem, the PET/iPET framework achieves the same levels of accuracy we obtain when we train with the full dataset using only 100 labeled examples. We attribute this to the fact that covertly aggression is very subtle, even the classification model with all the labeled examples struggles to detect this class.

As we increase the training set size, the performance gains of PET and iPET become smaller, but for 10 to 100 training examples, these methodologies considerably outperform standard supervised training in all datasets.

## 6. Conclusions

No augmentation technique consistently improved performance for BERT models. In one dataset, synonym replacement provided evidence of performance improvement. Adjectives scales with Grayscaling is an area that should be explored further on larger and better labeled datasets. For time restrictions we did not explore how all considered augmentations techniques could work in combination. This could offer more diversity in the augmented examples leading to better results.

Few-shot learning experiments show consistent improvement over supervised training, and seem very promising when classes are easily separable. As we increase the training set size, the performance gains of PET and iPET become smaller, but for 10 to 100 training examples, these methodologies considerably outperform standard supervised training in all datasets. Further exploration and tests with other datasets would be valuable.

# Appendices

## A. Training Procedure Details

**Training procedure (TRAC-2):** We train each BERT model for up to 2-3 epochs (task A/task B), using batch sizes of 16. The learning rate is set to 5e-5 and we use categorical cross-entropy loss function. We set a maximum sequence length of 150 tokens. Only less than 1% of the training data have sequences longer than this.

**Training procedure (RtGender):** We train each BERT model for up to 2 epochs, using batch sizes of 8. The learning rate is set to 2e-5 and we use categorical cross-entropy loss function. We set a maximum sequence length of 287 tokens. Only less than 1% of the training data have sequences longer than this.

## B. PET Charts

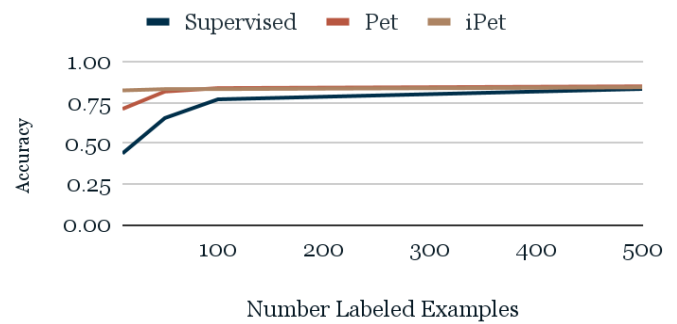

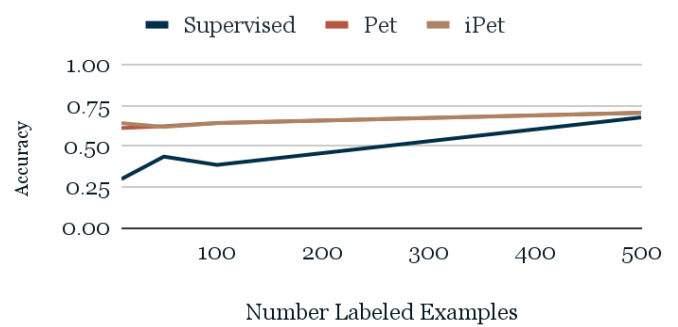

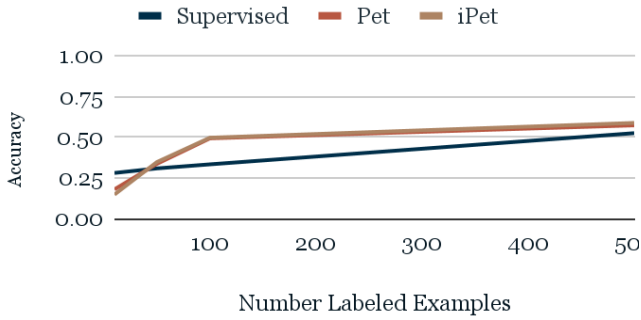

### C. Greyscaling

Using the mean of arc cosine as a measure of similarity among paragraph representations, we explore strong/weak matches.

**RtGender:** we find a high level of contextually appropriate augmented examples. 75% of the overall data have a score of 98% or higher. This drops down to 92% for the filtered dataset (excluding 'mixed' sentiment and irrelevant examples).

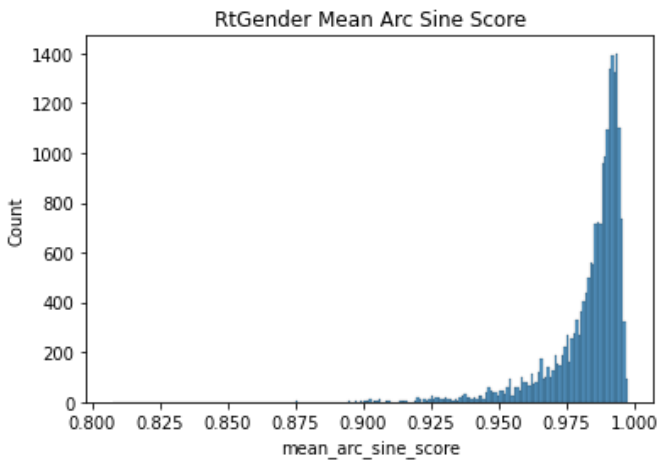

| Original | Augmented | Mean Arc Sine Score |
|---|---|---|
| **Poor Matches** | | |
| Good, luck,Shelby!!!! | adequate, luck,Shelby!!!! | 0.846 |
| Good, luck,Shelby!!!! | sufficient, luck,Shelby!!!! | 0.842 |
| YOURE, NOT, MY, REAL, DAD! | YOURE, NOT, MY, good, DAD! | 0.923 |
| **Strong Matches** | | |
| An amazing talk, with some **awesome** footage. I especially liked the two speakers in the film: the young girl made me feel humble, and the old man inspired me to take a step back and appreciate my luck. | An amazing talk, with some **good** footage. I especially liked the two speakers in the film: the young girl made me feel humble, and the old man inspired me to take a step back and appreciate my luck. | 0.994 |
| | ...**great**... | 0.995 |

TABLE 7. EXAMPLES OF POOR REPRESENTATIONS

**TRAC-2 Task A:** For this task, Greyscaling achieved second best results. The distribution of the similarity scores is wider for this dataset vs overall RtGender (75% of the data have scores of 93% or higher).

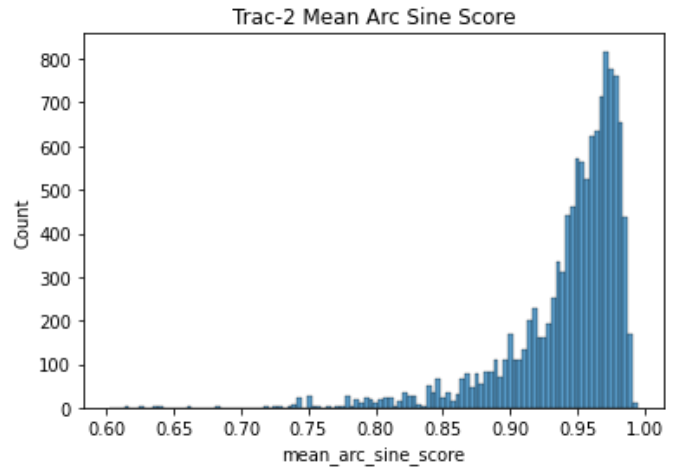

### D. RtGender Performance by Class (w/o Irrelevant examples)

The worst performing class using F1 scores is 'mixed.' The BOW model F1 'mixed' score was barely above 10%. Improvement from a BOW model to a transformer model led to substantial increases across categories and reduced the differences in scores for the overall vs overall excluding irrelevant examples.

| Model + Augmentation | Overall / Excluding Irrelevant | | | | | | | |
|---|---|---|---|---|---|---|---|---|
| | Pos. | | Neutral | | Mixed | | Neg. | |
| | Ove. | Exc. Irr. | Ove. | Exc. Irr. | Ove. | Exc. Irr. | Ove. | Exc. Irr. |
| BoW | 0.743 | 0.687 | 0.446 | 0.249 | 0.106 | 0.122 | 0.43 | 0.254 |
| BERT | **0.825** | **0.837** | **0.585** | 0.527 | 0.253 | 0.236 | **0.592** | **0.59** |
| BERT + Oversampling | **0.825** | 0.831 | **0.585** | 0.505 | 0.253 | **0.274** | **0.592** | 0.563 |
| BERT + EDA (+ O) | 0.81 | 0.815 | 0.578 | 0.578 | 0.201 | 0.241 | 0.59 | 0.587 |
| BERT + Back transl. (+ O) | 0.82 | 0.791 | 0.545 | 0.545 | 0.261 | 0.215 | 0.553 | 0.558 |
| BERT + Greysc. (+ O) | 0.798 | 0.824 | 0.556 | 0.529 | **0.257** | 0.232 | 0.553 | 0.565 |
| BERT + Greysc. + C.R. (+ O) | 0.808 | 0.811 | 0.521 | **0.581** | 0.236 | 0.231 | 0.559 | 0.571 |

TABLE 8. RtGender F1 Scores by Class

Given the wide discrepancy in labeling rates, we reviewed the labeling methodology. Researchers crowd-sourced labeling using Amazon's Mechanical Turk paying $0.20 for a run (5 post-response pairs). According to Voigt et al., the RtGender dataset annotations included attention controls. But, there were not multiple reviewers for each annotation.

**'Mixed' sentiment labeling may be flawed:** After reviewing labels ourselves, we believe that there may be inaccuracy in the 'mixed' sentiment bucket. We disagreed with the labeling for 'positive', 'negative', and 'neutral' less frequently.

| Mixed Sentiment Example | Our suggested classification |
|---|---|
| 'Just pointing out in the interests of avoiding misleading information, "mimeme" is not a Greek word. The Greek word is mimesis on which mimeme is formed then shortened to meme. What\'s more it does not mean "that which is imitated" but rather the thing that imitates or the imitation itself.' | Neutral |
| "It's a shame. He has conflated two talks - one on understanding medical stats and the other on stress management/premortem strategies. Both would have been interesting. Merging both feels like a muddle." | Negative |
| 'Fact. Source: Just read it...had chills' | Neutral |
| 'Some assholes might have been sexually harassing female childs after adopting them. Govt goes and do what it knows best. Every male is bad and no one will get it....hurr durr' | Neutral |
| 'Anyway, tell Orly she meant "cronies."' | Mixed |

TABLE 9. RtGender Random Sample of Mixed (ex. Irrelevant responses) & Our Suggested Rating